\documentclass[sn-mathphys-num]{sn-jnl}% Math and Physical Sciences Numbered Reference Style 
\usepackage{graphicx}%
\usepackage{multirow}%
\usepackage{amsmath,amssymb,amsfonts}%
\usepackage{amsthm}%
\usepackage{mathrsfs}%
\usepackage[title]{appendix}%
\usepackage{xcolor}%
\usepackage{textcomp}%
\usepackage{manyfoot}%
\usepackage{booktabs}%
\usepackage{algorithm}%
\usepackage{algorithmicx}%
\usepackage{algpseudocode}%
\usepackage{listings}%
\usepackage{colortbl}%
\usepackage{amsmath,amsfonts}%
\usepackage{makecell}
\theoremstyle{thmstyleone}%
\theoremstyle{thmstyletwo}%

\theoremstyle{thmstylethree}%

\begin{document}

\title[Article Title]{ReCLIP++: Learn to Rectify the Bias of CLIP for Unsupervised Semantic Segmentation}


\author[1]{\fnm{Jingyun} \sur{Wang}}\email{wangjingyun0730@gmail.com}

\author*[1]{\fnm{Guoliang} \sur{Kang}}\email{kgl.prml@gmail.com}

\affil[1]{\orgname{Beihang University}, \orgaddress{\city{Beijing}, \postcode{100190}, \country{China}}}

\abstract{
Recent works utilize CLIP to perform the challenging unsupervised semantic segmentation task where only images without annotations are available. 
However, we observe that when adopting CLIP to such a pixel-level understanding task, unexpected bias (including class-preference bias and space-preference bias) occurs. 
Previous works don't explicitly model the bias, which largely constrains the segmentation performance.
In this paper, we propose to explicitly model and rectify the bias existing in CLIP to facilitate the unsupervised semantic segmentation task. 
Specifically, we design a learnable ``Reference'' prompt to encode class-preference bias and a projection of the positional embedding in the vision transformer to encode space-preference bias respectively. 
To avoid interference, two kinds of biases are firstly independently encoded into different features, \emph{i.e.,} the Reference feature and the positional feature. 
Via a matrix multiplication between the Reference feature and the positional feature, a bias logit map is generated to explicitly represent two kinds of biases.
Then we rectify the logits of CLIP via a simple element-wise subtraction. 
To make the rectified results smoother and more contextual, we design a mask decoder which takes the feature of CLIP and the rectified logits as input and outputs a rectified segmentation mask with the help of Gumbel-Softmax operation.
A contrastive loss based on the masked visual features and the text features of different classes is imposed, which makes the bias modeling and rectification process meaningful and effective.
Extensive experiments on various benchmarks including PASCAL VOC, PASCAL Context, ADE20K, Cityscapes, and COCO Stuff demonstrate that our method performs favorably against previous state-of-the-arts. The implementation is available at: \url{https://github.com/dogehhh/ReCLIP}.
}




\maketitle

\section{Introduction}\label{sec1}

Semantic segmentation aims to attach a semantic label to each pixel of an image. 
Since the rise of deep learning~\cite{PLCA, kang2020contrastive,llmformer, yang2024multi, OV-VIS, liu2025m2ist}, semantic segmentation has been widely adopted in real-world applications, \emph{e.g.}, autonomous driving, medical image segmentation, \emph{etc}. 
Conventional approaches \cite{PSPNet, ECA-Net, MaskFormer, Mask2Former} for semantic segmentation have achieved remarkable performance.
However, the superior performance of those methods relies heavily on large amounts of fully annotated masks. Collecting such high-quality pixel-level annotations can be both time-consuming and expensive, \emph{e.g.}, some annotations for specialized tasks require massive expert knowledge, some are even inaccessible due to privacy reasons, \emph{etc}. Therefore, it is necessary to explore unsupervised semantic segmentation where only images without annotations are available.

% previous works
Unsupervised semantic segmentation (USS) has been studied for years. 
Many non-language-guided USS methods have been proposed, \emph{e.g.}, clustering-based methods~\cite{IIC, PiCIE, HSG, DeepSpectralMethods, autoregressive, ACSeg}, 
contrastive-learning-based methods~\cite{MaskContrast, MoCov2} and boundary-based methods~\cite{SegSort}, \emph{etc}.
Despite promising progress achieved, there still exhibits a large performance gap between USS and the supervised segmentation methods.
Besides, these methods typically obtain class-agnostic masks and have to depend on additional processing (\emph{e.g.}, Hungarian matching) to assign semantic labels to the masks, rendering them less practical in real scenarios. 

Recently, large-scale visual-language pre-trained models, \emph{e.g.}, CLIP \cite{CLIP}, demonstrate superior zero-shot classification performance by comparing the degree of alignment between the image feature and text features of different categories. 
A few CLIP-based USS approaches \cite{MaskCLIP, ReCo, CLIPpy, CLIP-S4, ReCLIP} emerge and show remarkable performance improvement compared with the non-language-guided USS methods. 
These models require no access to any types of annotations, and can directly assign a label to each pixel, benefiting from the aligned vision and text embedding space of CLIP.
However, good alignment between image-level visual feature and textual feature doesn't necessarily mean good alignment between pixel-level visual feature and textual feature. 
Thus, for CLIP, unexpected bias may inevitably appear.
Previous works don't explicitly model such bias, which may largely constrain their segmentation performance.

\begin{figure*}
  \centering
  \includegraphics[width=1\linewidth]{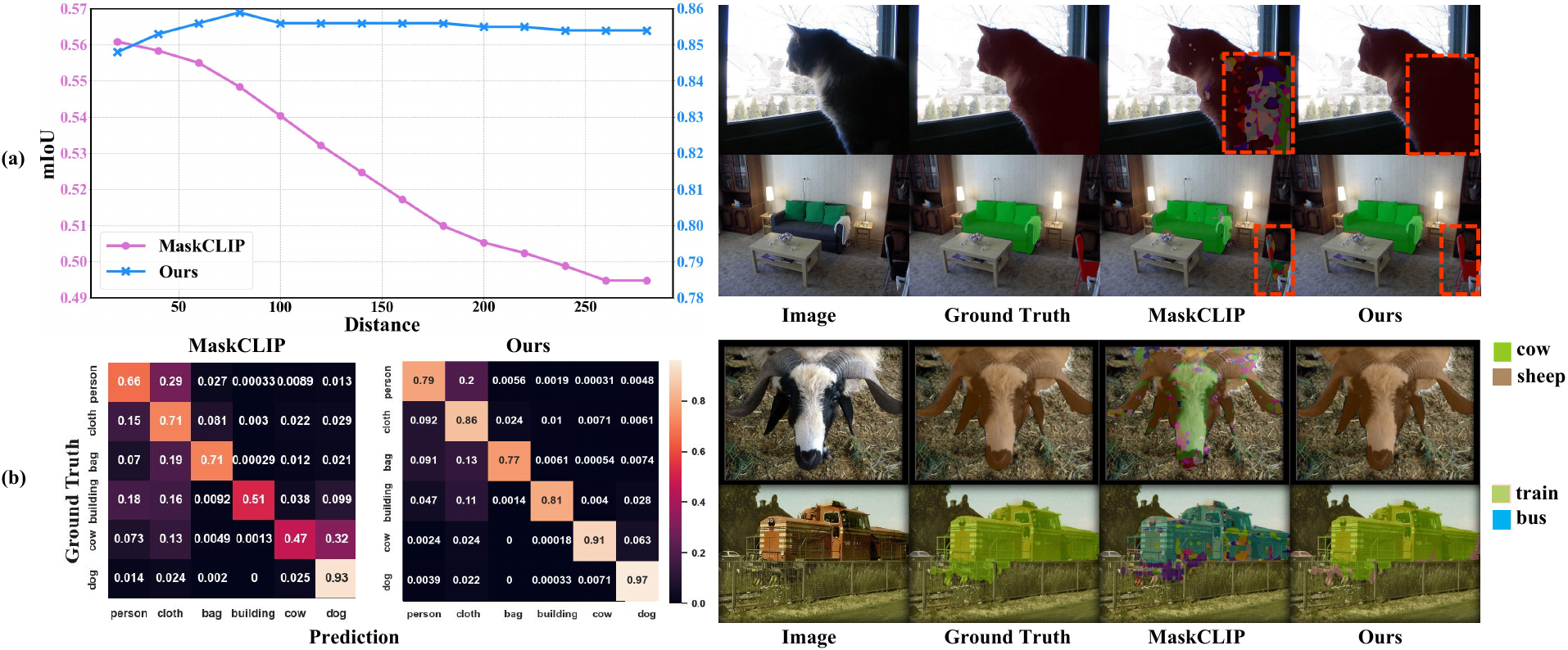}
  \caption{(a) \textbf{Space-preference bias.} 
  (Left): The relationship between distance ($x$-axis) and mIoU ($y$-axis) is drawn on PASCAL VOC \cite{PASCAL_VOC}. 
  The distance means the spatial distance between the object centroid and the image centroid and mIoU is computed based on predictions and ground truth.
  The curve shows that MaskCLIP~\cite{MaskCLIP} (pink) is apparently better at segmentation for central objects than boundary ones, but our method ReCLIP++ (blue) effectively mitigates this bias. 
  More details about how we draw this figure have been shown in our appendix.
  (Right): Visualizations also show our improvement in space-preference bias qualitatively.
  (b) \textbf{Class-preference bias.} 
  (Left): We randomly select 6 classes from PASCAL Context~\cite{PASCAL_Context} and draw the confusion matrix of MaskCLIP and our model.
  It shows that besides the ground truth, MaskCLIP also prefers to assign an incorrect but semantically relevant label to a pixel in quite a few cases, while our results show an apparent improvement. 
  (Right): The visualizations are consistent with what we observed in the confusion matrix. 
  For example, for a ``sheep'' (dark yellow), MaskCLIP tends to classify it as a ``cow'' (green) incorrectly.}
\label{introduction_fig}
\end{figure*}

When directly adopting CLIP in USS like MaskCLIP~\cite{MaskCLIP}, we observe two kinds of biases existing in CLIP.
From one aspect, as shown in Fig.~\ref{introduction_fig}~(a), CLIP exhibits space-preference bias. 
CLIP performs apparently better for segmenting central objects than the ones distributed near the image boundary. 
It can be reflected by the fact that mIoU decreases as the distance between the object centroid and the image centroid increases.
From the other aspect, as shown in Fig.~\ref{introduction_fig}~(b), there exists class-preference bias between semantically relevant categories in CLIP.
For example, according to visualizations (right), when the ground truth is a ``sheep'' (dark yellow), CLIP tends to incorrectly classify it as a ``cow'' (green). 
We also show such a trend between randomly selected classes by confusion matrix (left).
Elements on the diagonal line represent the right classification, while others are false. 
We observe that besides ground truth, CLIP usually prefers to assign an incorrect but semantically relevant label to a pixel in quite a few cases, exhibiting a wide range of class-preference bias of CLIP. 

In this paper, we propose to explicitly model and rectify the bias of CLIP to facilitate the USS. 
Specifically, we design two kinds of text inputs for each class, which are named ``Query'' and ``Reference'' respectively.
The Query is manually designed and fixed while the Reference contains learnable prompts.
We pass the Query and Reference through the text encoder of CLIP to obtain the Query feature and Reference feature.
We adopt the Query feature as the segmentation head to generate the Query logits for each pixel of the image, which represents the original segmentation ability of vanilla CLIP.
In contrast, we utilize the Reference feature to encode the class-preference bias.
In addition, we project the positional embedding of CLIP's vision transformer to generate positional feature, which encodes the space-preference bias.
The encoding processes of two biases are independent to avoid interference.
Via a matrix multiplication between the Reference feature and the positional feature, a bias logit map is generated to explicitly represent two kinds of biases.
Then, we remove the bias from the original CLIP via a logit-subtraction mechanism, \emph{i.e.,} subtracting the bias logits from the Query logits.
To make the rectified results smoother and more contextual, we concatenate the rectified logit map to the feature extracted by CLIP's visual encoder and pass them through a designed decoder. 
Then we apply the Gumbel-Softmax operation to the output of the decoder to generate a rectified mask.
To make the bias modeling and rectification process meaningful and effective, 
the contrastive loss based on masked visual features (\emph{i.e.,} applying rectified masks to the visual feature of CLIP and pooling) and text features of different categories is imposed.

We conduct extensive experiments on five standard semantic segmentation benchmarks, 
including PASCAL VOC~\cite{PASCAL_VOC}, PASCAL Context~\cite{PASCAL_Context}, ADE20K~\cite{ADE20K}, Cityscapes~\cite{cityscapes} and COCO Stuff~\cite{cocostuff}.
Experiment results demonstrate that ReCLIP++ performs favorably against previous state-of-the-arts.
Notably, on PASCAL VOC, our method outperforms MaskCLIP+~\cite{MaskCLIP} by $15.4$\% and CLIP-S4~\cite{CLIP-S4} by $13.4$\% mIoU. 
Extensive ablation studies verify the effectiveness of each design in our framework.

This paper is an extension of our conference paper~\cite{ReCLIP}.
To differentiate, we denote the method proposed in our conference version as ReCLIP and that in this paper as ReCLIP++.
%The key technical improvements of ReCLIP++ beyond ReCLIP are summarized as follows:
Compared to our conference version, we make further contributions, which are summarized as follows:
1) We optimize the design of the Bias Extraction Module. Specifically, we independently encode class-preference bias and space-preference bias into class-level Reference feature and patch-level positional feature respectively.
Then the bias logit map, which explicitly represents both biases, is obtained by combining two features with matrix multiplication.
2) We additionally introduce a mask decoder, which takes the rectified logit map and the visual feature of CLIP as input and outputs smoother and more contextual rectified predictions.
3) We design a new strategy to generate a more accurate multi-label hypothesis for each image, which provides better supervision for the bias rectification process.
4) Benefiting from our new design, the distillation stage in ReCLIP is no longer necessary. We remove the distillation stage to simplify training and achieve even better segmentation performance. % and higher efficiency with fewer training epochs.}
5) We evaluate our methods on two more datasets including Cityscapes~\cite{cityscapes} and COCO Stuff~\cite{cocostuff}.
On all the datasets, the segmentation performance of ReCLIP++ outperforms  ReCLIP remarkably.

In a nutshell, our contributions are summarized as follows:
\begin{itemize}
\item 
We observe that when applying CLIP to pixel-level understanding tasks, unexpected biases, including space-preference bias and class-preference bias, occur. 
These biases may largely constrain the segmentation performance of CLIP-based segmentation models.
\item 
We propose to explicitly encode the class-preference and space-preference bias of CLIP via learnable Reference text inputs and projection of positional embedding respectively, and model both biases into one bias logit map via matrix multiplication. 
Through a simple logit-subtraction mechanism and the contrastive loss built on masked features of different classes, we effectively rectify the bias of CLIP. 
\item 
We conduct extensive experiments on segmentation benchmarks under the USS setting. Experiment results show superior performance of our method to previous state-of-the-arts. 
\end{itemize}

\section{Related Work}\label{sec2}

\noindent \textbf{Pre-trained vision-language models}.
Pre-trained vision-language models (VLMs)~\cite{Uniter, Virtex, Unicoder-vl, AlignBeforeUse, Hero} have developed rapidly, largely due to the availability of large-scale image-text pairs available on the Internet. 
Early VLMs typically rely on joint embedding spaces and multi-modal transformers to align visual and textual information.
They often use objectives such as masked language modeling, image-text matching, and cross-modal retrieval.
More recently, contrastive learning-based models such as CLIP~\cite{CLIP}, ALIGN~\cite{ALIGN}, and SLIP~\cite{Slip} have achieved remarkable progress by training on hundreds of millions of image-text pairs. 
These models employ a dual-encoder architecture, where images and texts are encoded separately, and a contrastive loss is used to maximize the similarity of matched image-text pairs while minimizing the similarity of mismatched pairs. 
This approach enables efficient, large-scale training and robust alignment at the image level.
As a result, pre-trained VLMs demonstrate strong zero-shot transfer capabilities, excelling in tasks such as zero-shot classification, and can be adapted to various downstream applications, including object detection~\cite{DetPro, UniDetector} and semantic segmentation~\cite{MaskCLIP,tcl, SAN}.

\noindent \textbf{Unsupervised semantic segmentation.} 
While conventional approaches of semantic segmentation~\cite{ECA-Net, PSPNet} rely on pixel-level annotations and weakly-supervised methods~\cite{WSSS1, WSSS2, WSSS3,c2am} still ask for image-level labels, unsupervised semantic segmentation (USS) methods~\cite{PiCIE, scan, autoregressive, MaskContrast, STEGO} explore to train a segmentation model without any annotations. 
Models like~\cite{GAN1, GAN2} adopt generative model~\cite{GAN} to separate foreground with background or generate corresponding masks.
SegSort~\cite{SegSort}, HSG\cite{HSG} and ACSeg~\cite{ACSeg} use clustering strategy, while
IIC~\cite{IIC} uses mutual information maximization to perform unsupervised learning.
MaskContrast~\cite{MaskContrast} introduces contrastive learning into USS. 
Others like DSM~\cite{DeepSpectralMethods} and LNE~\cite{LNE} exploit features extracted from self-supervised models and spectral graph theory to facilitate segmentation. 
However, the methods mentioned above either fail to segment images with multi-category objects or show a large performance gap with the supervised methods. Besides, they can only obtain class-agnostic masks and have to depend on additional strategies, such as Hungarian-matching algorithm~\cite{Hungarian}, to match the corresponding category with the segment.
Recently, pre-trained vision-language models are adopted in USS. 
MaskCLIP~\cite{MaskCLIP} modifies the image encoder of CLIP to generate patch-level features and directly performs segmentation with text features as classifiers. 
CLIP-py~\cite{CLIPpy} performs contrastive learning between visual features from self-supervised ViT~\cite{Self-supervised} and text features from CLIP. 
ReCo~\cite{ReCo} performs image retrieval with CLIP and extracts class-wise embedding as classifier with co-segmentation. 
CLIP-S4~\cite{CLIP-S4} learns pixel embeddings with pixel-segment contrastive learning and aligns such embeddings with CLIP in terms of embedding and semantic consistency. These methods can directly assign a label to each pixel, which falls into the category of language-guided unsupervised semantic segmentation.
However, directly applying CLIP in USS may result in unexpected bias. 
No previous method considers such bias. 
In this paper, we propose to explicitly model and rectify the bias of CLIP for unsupervised semantic segmentation.

\noindent \textbf{Language-guided semantic segmentation.}
Recently, many works explore semantic segmentation guided by language under different settings.
Zero-shot works~\cite{ZS3Net, SPNet, STRICT, LSeg} split classes into seen and unseen sets. During the training period, only masks of seen classes are provided. For inference, models are tested on both seen and unseen classes, but the test data is still in the same distribution as the training data.
Trainable open-vocabulary works~\cite{GroupViT, OpenSeg, OVSegmentor, SAN, ViewCo, OVSeg, tcl, CoCu} are trained in one scenario with extra annotations including class-agnostic masks or image captions but are used for predicting segmentation masks of novel classes in other scenarios. 
Recent training-free open-vocabulary works~\cite{ClearCLIP, SCLIP, MaskCLIP} adapt CLIP for semantic segmentation by modifying its architecture.
These methods can directly perform inference on downstream datasets without any training.
From the technical view, our method also falls into the category of language-guided semantic segmentation. 
However, we consider the unsupervised setting. 
In this setting, we have access to images without any annotations during training.
The training and inference images are sampled from the same distributions and the same set of categories.
Such a setting is different from the typical zero-shot or trainable open-vocabulary setting.
% {TODO: a brief discussion about training-free OVSS?}

\section{Method}\label{sec3}

\noindent \textbf{Background} In this work, we aim to rectify the bias of CLIP for unsupervised semantic segmentation. 
In USS, we only have access to images without any types of annotations to train the segmentation model.
For training and inference, the same set of categories are considered and the data distributions are assumed to be the same.

\noindent \textbf{Overview} 
We aim to rectify the bias of CLIP including the class-preference bias and the space-preference bias, to facilitate unsupervised semantic segmentation.
From a high level, class-preference bias reflects the shift of CLIP predictions towards specific classes, while space-preference bias reflects the shift of CLIP predictions towards specific spatial locations. Both biases will be finally reflected in the bias logit map. A reasonable way to rectify the bias is to subtract the bias logit map from the query logit map predicted by original CLIP.

\begin{figure*}
  \centering
  \includegraphics[width=1\linewidth]{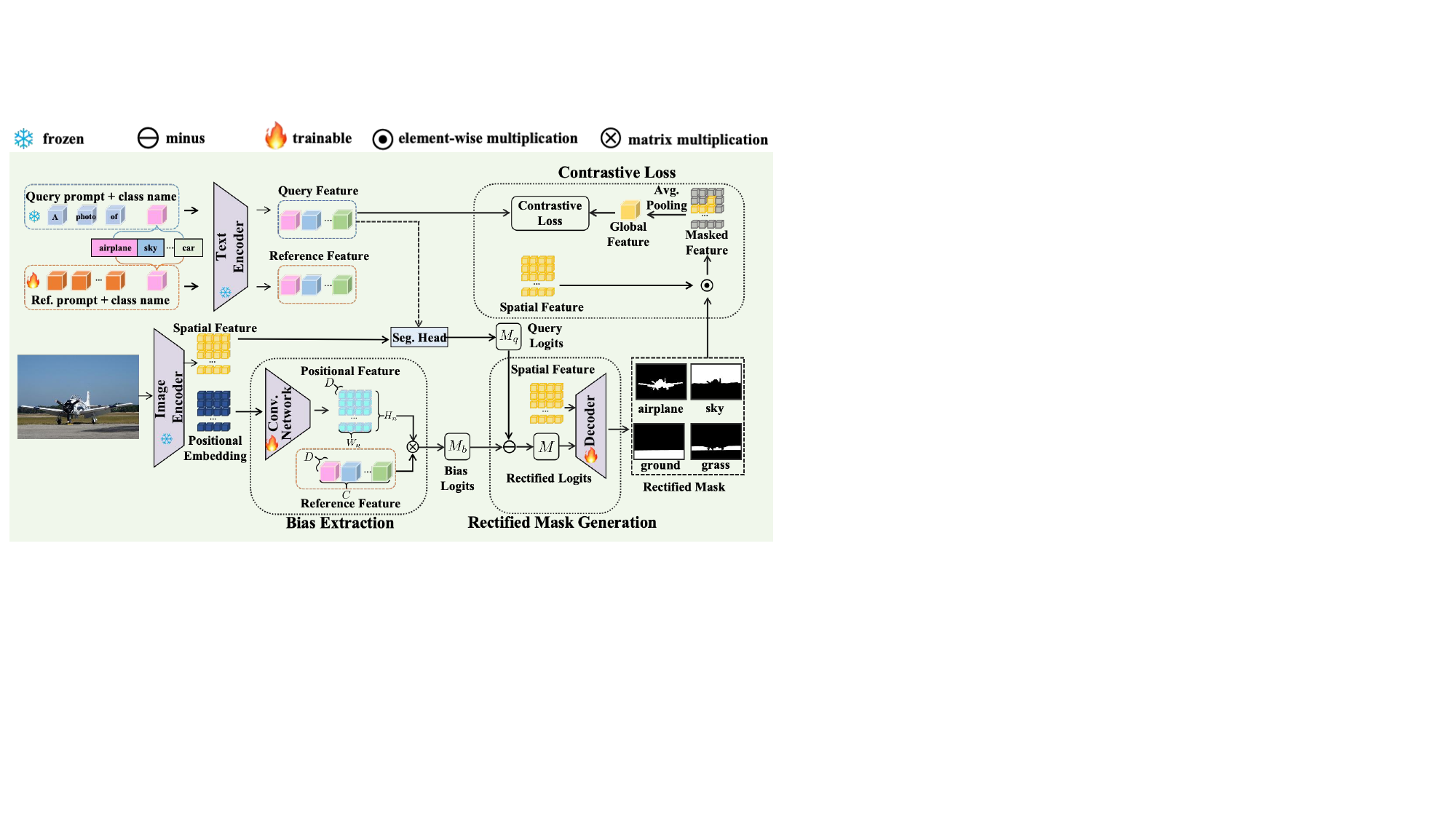}
  \caption{\textbf{Method overview of ReCLIP++.}
   We propose a new framework for language-guided unsupervised semantic segmentation. We aim to rectify the class-preference and space-preference bias of CLIP via specifically designed Bias Extraction module, Rectified Mask Generation module, and Contrastive Loss module.
  }
\label{method_overview}
\end{figure*}

The general framework of our method is illustrated in Fig.~\ref{method_overview}. 
We first forward the image $I\in\mathbb{R}^{3 \times H\times W}$ through the image encoder of CLIP to obtain the patch-level image feature $Z$.
For each class, we manually design the text input which is named Query $Q$, and fix $Q$ throughout the training. 
We pass $Q$ through the text encoder of CLIP to obtain query text feature $W_q$ for each class, which serves as the weight of the query segmentation head. With $W_q$, we can obtain a query logit map $M_q$, which represents the segmentation ability of the original CLIP (Sec.~\ref{3.1}).

We extract the bias existing in CLIP in a learnable way. Specifically, we design a kind of learnable text input named Reference $R$ for each class. Passing $R$ through the text encoder of CLIP, we obtain Reference text feature $W_r$, which is expected to encode the class-preference bias.
Meanwhile, the positional embedding is projected into the positional feature $W_p$ to encode the space-preference bias.
Then the bias logit map $M_b$ is extracted via a matrix multiplication between $W_r$ and $W_p$ (Sec.~\ref{3.2}).

The Query logit map and the bias logit map are then passed through a Rectified Mask Generation module to generate rectified semantic masks. Firstly, a rectified logit map $M$ is generated via a simple subtraction operation between $M_q$ and $M_b$. Then, a mask decoder takes $M$ and visual feature $Z$ as input and outputs smoother and more contextual semantic masks with the help of Gumbel-Softmax operation (Sec.~\ref{3.3}).

To enable a more meaningful and effective bias rectification, we generate a multi-label hypothesis for each image and impose contrastive loss based on masked visual features and Query text features of different classes (Sec.~\ref{3.4}). 
CLIP is kept frozen during the training process.

\subsection{Baseline: Directly Segment with CLIP}
\label{3.1}
Following MaskCLIP~\cite{MaskCLIP}, we adapt pre-trained CLIP~\cite{CLIP} (ViT-B/16) to the semantic segmentation task. 
We remove the query and key embedding layers of last attention but reformulate the value embedding layer and the last linear projection layer into two respective $1\times 1$ convolutional layers.
Therefore, the image encoder can not only generate the patch-level visual feature for dense prediction but also keep the visual-language association in CLIP by freezing its pre-trained weights. We forward image $I$ through the image encoder and obtain patch-level feature $Z \in \mathbb{R}^{n \times D}$ ($n$ is the number of patches).

Each text $Q_i$ in Query $Q=\{Q_1, Q_2, \cdots, Q_C\}$ ($C$ is the number of classes) is an ensembling of several manually designed templates, \emph{e.g.}, ``a good/large/bad photo of a $[CLS]$", where $[CLS]$ denotes a specific class name. 
Passing $Q_i$ through the text encoder, we obtain different text embeddings for different manually designed templates. We average all those text embeddings as the embedding of $Q_i$.
Then the embeddings of all the $Q_i$ in $Q$ make up the query text feature $W_q\in \mathbb{R}^{C\times D}$.
We treat text feature $W_q$ as the weight of the segmentation head to perform 1 $\times$ 1 convolution. By sending feature $Z$ to the segmentation head, we get a Query logit map $M_q \in \mathbb{R}^{n\times C}$. 
Then the segmentation mask can be predicted by $\arg\max$ operation on $M_q$ across the class channel $C$. 

\subsection{Bias Extraction}
\label{3.2}
In the Bias Extraction Module, we aim to encode the class-preference bias and the space-preference bias of CLIP independently.
Then we combine them into a bias logit map to explicitly model both biases.

\textbf{Class-preference Bias Encoding.}
In order to encode the class-preference bias brought by pre-trained CLIP with respect to a specific segmentation task, we design learnable Reference text $R_i$ in $R=\{R_1, R_2, \cdots, R_C\}$ as an additional text input for each class. 
Inspired by CoOp~\cite{coop}, $R_i$ consists of a learnable prompt which is shared across all the classes for efficiency, and a class name $[CLS]$, which can be formed as 
\begin{equation}
R_i = [v_1][v_2]...[v_l]...[v_{L}][CLS] \label{reference prompt},
\end{equation}
where each $[v_l] (l \in \{1,...,L\})$ is a vector with dimension $D$ and serves as a word embedding. 
The $L$ is a constant representing the number of word embeddings to learn. 
Totally, there are $77$ word embeddings for a textual input of CLIP. 
Among these, two word embeddings are used for representing class names and two for indicating the start and end of a textual input. Thus, we set $L$ to $73$. 

Passing Reference $R$ through the text encoder, we obtain reference text feature $W_r \in \mathbb{R}^{C\times D}$, which is expected to encode the class-preference bias.
As Reference is only class-aware, the Reference feature is less likely to encode any space-related information.

\textbf{Space-preference Bias Encoding.}
In ViT~\cite{ViT}, positional embedding (PE) is important for encoding spatial information into features. 
Thus, we assume the space-preference bias should depend on the positional embedding (PE) and we choose to learn a projection of PE to encode the space-preference bias.
The projection network of PE is designed as a 1-layer $1\times1$ convolutional network.
Then we project PE $p$ by the designed convolutional network to obtain positional feature $W_p \in \mathbb{R}^{n\times D}$.
During the training process, the projection network is optimized to encode the space-preference bias with respect to each patch of the image. 
Note that the encoding process of space-preference bias is independent of that of class-preference bias because the encoding of space-preference bias doesn't rely on any class information.

\textbf{Bias Logits Generation.}
With Reference feature $W_r$ encoding the class-preference bias and positional feature $W_p$ encoding the space-preference bias, we combine the encoding results of two biases to generate the final bias logit map $M_b \in \mathbb{R}^{n\times C}$, via a matrix multiplication as
\begin{equation}
M_b = W_p \times W_r^T. \label{bias logits}
\end{equation}
Finally, $M_b$ represents the final bias that we aim to remove from the predicted query logit map $M_q$ of the original CLIP.

\textbf{Bias Extraction: ReCLIP++ vs. ReCLIP.}
Note that the way of bias extraction in ReCLIP++ is different from that in ReCLIP~\cite{ReCLIP}.
In ReCLIP, to extract the class-preference bias, we compute the similarity between Reference features of different classes and CLIP's visual feature map $Z$ to generate a class-preference bias logit map.
However, as CLIP's image encoder takes PE as input to encode spatial information into feature $Z$, the class-preference bias logit map may inevitably encode some space-related information. 
Consequently, the implicit space-related information in extracted class-preference bias may interfere with the extraction of space-preference bias, resulting in less effective bias extraction and rectification.

Differently, in ReCLIP++, we first encode class-preference bias and space-preference bias into the Reference feature and positional feature respectively. The reference feature only depends on class-related information (\emph{i.e.,} learnable Reference text), while the positional feature only depends on space-related information (\emph{i.e.,} PE).
Thus, in this way, we may \textit{independently} encode two kinds of biases and largely avoid the interference.
As a result, with ReCLIP++, we realize more effective bias rectification and achieve better segmentation results (see Sec.~\ref{sec4} for more details). 

\subsection{Rectified Mask Generation}
\label{3.3}
{
By keeping Query prompt $Q$ fixed, the Query logit map $M_q$ represents the natural prediction ability of the original CLIP model, which may contain class-preference bias and space-preference bias.
With the bias extraction module, we obtain $M_b$ which represents the bias that we aim to remove from CLIP predictions.
In this paper, we perform a simple element-wise subtraction between $M_q$ and $M_b$ to generate the rectified logit map $M$, \emph{i.e.,}
\begin{equation}
M = M_q - M_b. \label{minus_operation}
\end{equation}
From a high level, this operation can be interpreted as subtracting the ``bias" from the predictions with Query text feature $W_q$ serving as the weight of the segmentation head. 

Though the rectified logit map $M$ can be directly used to generate semantic masks, the local context is not fully considered, which may result in non-smooth mask predictions.
The reason is as follows.
The query logit map $M_q$ is computed as the cosine similarity between each patch feature and textual features, treating each patch independently without considering information from neighboring patches.
The bias logit map $M_b$ is generated from the reference feature $W_r$ (which encodes only class-related information and is spatially invariant) and the positional feature $W_p$ (which is encoded with $1\times 1$ convolution and does not contain semantic information), without considering spatial interaction.
Since both $M_q$ and $M_b$ are computed independently for each patch, the resulting rectified logit map $M$ is constructed in a patch-wise manner.
As a result, directly using the rectified logit map $M$ for mask prediction may result in non-smooth masks because each patch is predicted without awareness of its local neighbors.

Thus, different from our conference version~\cite{ReCLIP}, we additionally introduce a mask decoder $\mathcal{F}_{dec}$ to transform $M$ into a smoother and more contextual logit map. 
In our implementation, $\mathcal{F}_{dec}$ consists of a convolutional layer with a  $5 \times 5$ kernel and a batch normalization layer.
We concatenate the rectified logit map $M$ to the visual feature $Z$ of CLIP and pass them through the mask decoder $\mathcal{F}_{dec}$ to generate the rectified output $M_o\in\mathbb{R}^{n \times C}$, \emph{i.e.,}
\begin{equation}
M_o = \mathcal{F}_{dec}(M, Z).
\label{eq-mo}
\end{equation}

To generate the semantic mask $\hat{M}\in\mathbb{R}^{n\times C}$ for each category, a natural way is to employ $\arg\max$ operation to $M_o$ across the class channel $C$ and generate the one-hot label for each spatial location.
Then, each channel of $\hat{M}$ corresponds to the mask for a specific category.
However, as the $\arg\max$ operation is not differentiable, we are not able to optimize $\hat{M}$ end-to-end through gradient back-propagation during the training process. 
Thus we utilize a Gumbel-Softmax trick~\cite{gumbel_softmax} to generate the semantic mask for each category with $\tau_1$ as the temperature, \emph{i.e.,}
\begin{equation}
\hat{M} = \text{Gumbel-Softmax}(M_o, \tau_1).
\end{equation} 
In this way, with our designed supervision (Sec.~\ref{3.4}), 
the gradient can be back-propagated through $\hat{M}$ to encourage a more effective bias rectification and a better semantic mask generation.

\subsection{Bias Rectification via Contrastive Loss}
\label{3.4}
In this section, we introduce a contrastive loss to supervise the rectification process.

\textbf{Image-level multi-label hypothesis generation.}
In order to compute the contrastive loss, we need to know what classes exist in a specific image.
However, this information is not available under the USS setting. 
In this section, we propose to generate an image-level multi-label hypothesis which means the set of classes that potentially exist in an image, to facilitate the following contrastive loss calculation.

In ReCLIP~\cite{ReCLIP}, we directly calculate the similarity scores between the class-level text features of manually designed prompts, \emph{e.g.}, ``a photo of a $[CLS]$", and the image-level visual feature extracted by CLIP. 
We select the set of classes whose scores are higher than a threshold as the hypothesis.
However, the objects in an image may have different scales (large/small) and may be located arbitrarily (center/boundary).
As CLIP is pre-trained to align image-level visual features with text features, it may focus on the most salient objects in an image and fail to recognize every existing object.
Thus, generating the multi-label hypothesis based on the \textit{image-level} visual features (like ReCLIP) cannot fully discover the underlying categories in an image, especially for categories whose objects are usually small or less salient in an image.

In ReCLIP++, we design a new strategy to generate a more accurate multi-label hypothesis for an image.
Given an image, we first split it into overlapped crops. 
Specifically, we slide a window across an image with a certain stride.
In our implementation, we set the window width and height to $r$ times the width and height of the input image and set the horizontal and vertical stride to half of the window width and height respectively.
For each crop, we forward it through the visual encoder of CLIP to obtain its global visual feature and compute the similarity scores between the global visual feature and the Query feature $W_q$.
We treat the class with the highest score as the detected class for this crop. 
Then for a given image, among all of its crops, we calculate the frequency $f(k)$ of the $k$-th class, $k\in\{1, 2, ..., C\}$, being detected.
We choose the classes with frequency higher than the threshold $t$ as the multi-label hypothesis $\mathcal{H}$ for an image, \emph{i.e.,}
\begin{equation}
\mathcal{H} = \{k|k \in\{1, 2, ..., C\} \text{ and } f(k) > t\}.
\label{hypothesis}
\end{equation}
The rationale behind the strategy of multi-label hypothesis in ReCLIP++ is that the objects that are not salient globally may be salient locally. 
By aggregating the predictions from multiple local regions, ReCLIP++ can better discover existing categories in an image, especially for those that are small or less salient. 

\textbf{Bias Rectification via Contrastive Loss.}
We then design a contrastive loss to supervise the rectification process via aligning masked features with text features of corresponding classes.
Specifically, we expand $\hat{M}$ to $\hat{M'} \in \mathbb{R}^{n\times C\times D}$ and $Z$ to $Z' \in \mathbb{R}^{n\times C\times D}$, apply $\hat{M'}$ to the feature $Z'$, and perform a global average pooling to get the class-level masked features $Z_g\in\mathbb{R}^{C\times D}$, which encode the features of regions belonging to different classes:
\begin{equation}
Z_g = \text{Avg-Pool}(\hat{M'} \otimes Z'),
\end{equation}
where $\otimes$ represents element-wise multiplication.

We compute the similarities between $Z_g$ and the text features of all the categories.
As the masked features $Z_g$ represent the objects of interest without context, we utilize $W_q$ generated by text input $Q$ for similarity computation.
With generated $\mathcal{H}$, the contrastive loss is computed as
\begin{equation}
%\begin{split}
\mathcal{L} = -\frac{1}{|\mathcal{H}|} \sum_{k\in\mathcal{H}}\log\frac{\exp\{S_{k,k}/\tau\}}{\sum_{j=1}^C\exp\{S_{k,j}/\tau\}} \label{contrastive-loss} 
%\end{split}
\end{equation}
where $S_{k,j}$ denotes the cosine similarity between visual feature of class $k\in\mathcal{H}$ (denoted as $Z_{g,k}$) and the text feature of $j$-th ($j\in \{1, 2, \cdots, C\}$ category (denoted as $W_{q,j}$) and the $\tau$ is a constant.
Among it, $S_{k,j}$ can be calculated as follows ($<\cdot,\cdot>$ means dot product):
\begin{equation}
S_{k,j} = \frac{<Z_{g, k}, W_{q,j}>}{\|Z_{g, k}\|\|W_{q,j}\|}.
\end{equation}

A better modeling of bias yields more accurate estimations of object masks. Then the masked features of objects are more aligned with the corresponding text features. As a result, the contrastive loss (Eq.~(\ref{contrastive-loss}) will therefore be lower.
In contrast, a worse modeling of bias results in higher contrastive loss.
Thus, minimizing Eq.~(\ref{contrastive-loss}) will drive the model to update towards making more accurate mask predictions, \emph{i.e.}, rectifying the bias of CLIP when adapting CLIP to the downstream USS task.

Note that in ReCLIP, an additional distillation stage is employed, where the knowledge of rectified CLIP is distilled into a specifically designed segmentation network like DeepLab-v2~\cite{Deeplabv2}.
However, in ReCLIP++, we don't need such a distillation stage. %as employed in ReCLIP. 
Empirically, we observe that such a distillation process has two opposing effects.
On the one hand, it may benefit the smoothness of the predicated mask, which has a positive effect on the segmentation performance.
On the other hand, instead of mitigating the bias, it may enlarge the bias to an extent, which has a negative effect on the segmentation performance.
Benefitting from $\mathcal{F}_{dec}$, the rectified mask of ReCLIP++ is much smoother and contextual than that in ReCLIP, rendering the negative effect of distillation outweigh the positive effect of distillation. 
Thus, we choose not to employ the distillation in ReCLIP++ to make the training more simplified and achieve even better segmentation performance (see Sec.~\ref{ablation_study} for more details).
%Empirically, we observe that instead of mitigating the bias, distillation process may enlarge the bias to an extent. Previously in ReCLIP, the distillation mainly benefits the smoothness of predicted masks, and thus improves the segmentation performance. However, in ReCLIP++, (See Sec.~\ref{} for more details).

\subsection{Inference} 
\label{3.5}
%Finally, we obtain a rectified CLIP.
For the inference with ReCLIP++, we obtain the query logit map $M_q$ by the query segmentation head %as Sec.~\ref{3.1} 
and the bias logit map $M_b$ which models both biases by the Bias Extraction Module. %as Sec.~\ref{3.2}.
We then obtain the rectified output $M_o$ by the Rectified Mask Generation Module with Eq.~(\ref{eq-mo}). % as Sec.~\ref{3.3}.
With a simple $\arg\max$ operation to $M_o$ across the class channel $C$, we generate the rectified masks as final predictions.

\section{Experiment}\label{sec4}

\subsection{Setup}
\label{implementation}
\noindent \textbf{Datasets.} 
We conduct experiments on five standard benchmarks for semantic segmentation, including PASCAL VOC 2012~\cite{PASCAL_VOC}, PASCAL Context~\cite{PASCAL_Context}, ADE20K~\cite{ADE20K}, Cityscapes~\cite{cityscapes} and COCO Stuff~\cite{cocostuff}. 
PASCAL VOC 2012~(1,464/1,449 train/validation) contains 20 object classes, 
while PASCAL Context~(4,998/5,105 train/validation) is an extension of PASCAL VOC 2010 and we consider 59 most common classes in our experiments. 
ADE20K~(20,210/2,000 train/validation) is a segmentation dataset with various scenes and 150 most common categories are considered. 
Cityscapes~(2,975/500 train/validation) consists of various urban scene images of 19 categories from 50 different cities.
COCO Stuff~(118,287/5,000 train/validation) has 171 low-level thing and stuff categories excluding background class. 
Following the previous evaluation protocol of unsupervised semantic segmentation methods~\cite{ReCo, STEGO}, we use 27 mid-level categories for training and inference. 

\noindent \textbf{Implementation details.}
For the image encoder of CLIP~\cite{CLIP}, we adopt ViT-B/16 as our visual backbone. 
%{The official code of MaskCLIP~\cite{MaskCLIP} adopts GeLU as its activation function in the image encoder, while we adopt QuickGeLU following vanilla CLIP for better performance.}
For the text encoder of CLIP, we adopt Transformer~\cite{Transformer}.
During the whole training period, we keep both of the encoders frozen. 
We use conventional data augmentations including random cropping and random flipping.
We use an SGD~\cite{SGD} optimizer with a learning rate of 0.01 and a weight decay of 0.0005. 
We adopt the poly strategy with the power of 0.9 as the learning rate schedule.
We set $r = 1/6$ for all the datasets. We set $t = 7\%$ for PASCAL VOC and Cityscapes, and $t = 0$ for the rest datasets.
In our experiment, we report the mean intersection over union (mIoU) as the evaluation metric.
More details about training can be found in our appendix.

\subsection{Comparison with previous state-of-the-arts}
\label{comparison}

\noindent \textbf{Baselines.}
We mainly compare our method with three types of semantic segmentation methods to verify the superiority of our method:
(1) Trainable language-guided methods for OVSS task (TL-OVSS), including GroupViT~\cite{GroupViT}, CoCu~\cite{CoCu}, and TCL~\cite{tcl};
(2) CLIP-based methods for training-free OVSS task (TF-OVSS), including CLIP~\cite{CLIP} (vanilla CLIP visual encoder), MaskCLIP~\cite{MaskCLIP} (CLIP visual encoder with modified last-attention block), SCLIP~\cite{SCLIP}, and ClearCLIP~\cite{ClearCLIP};
(3) CLIP-based methods for USS task (C-USS), including MaskCLIP+~\cite{MaskCLIP}, CLIPpy~\cite{CLIPpy}, ReCo~\cite{ReCo}, CLIP-S4~\cite{CLIP-S4} and ReCLIP~\cite{ReCLIP}.
We directly cite the corresponding results from the original papers, except that $\dag$ means the results are obtained by running the officially released source code and ${\ddagger}$ means the results are cited from TCL~\cite{tcl}.
All the numbers reported are presented as percentages.
Among these, TL-OVSS methods rely on weak annotations like image-caption pairs to train the model, while TF-OVSS methods can directly perform open-vocabulary segmentation without any training.
C-USS methods share the same language-guided unsupervised setting with ReCLIP++. 

\begin{table}[!t]
\setlength{\tabcolsep}{6pt}
\caption{
\textbf{Comparison with trainable language-guided methods for OVSS (TL-OVSS), CLIP-based methods for training-free OVSS (TF-OVSS), and CLIP-based methods for USS (C-USS) on five various benchmarks.}
{\dag} means the results are obtained by running the officially released source code and ${\ddagger}$ means the results are cited from TCL~\cite{tcl}.
\label{comparison_table}}
\centering
\renewcommand{\arraystretch}{1.3}
\begin{tabular}{c|c|c|ccccc}
\hline 
Method & Pub. & Setting & VOC & Context  & ADE & City & Stuff\\
\hline
GroupViT~\cite{GroupViT}$^{\ddagger}$& CVPR'22& \multirow{3}{*}{\makecell[c]{TL-OVSS}} &$79.7$& $23.4$& $9.2$& $11.1$&$15.3$\\
TCL~\cite{tcl}& CVPR'23 & &$77.5$ & $30.3$& $14.9$&$23.1$ &$19.6$\\
CoCu~\cite{CoCu}& NeurIPS'24& &-& -& $11.1$& $15.0$&$13.6$\\
\hline
CLIP~\cite{CLIP}$^\dag$& ICML'21&\multirow{4}{*}{\makecell[c]{TF-OVSS}}& $41.8$&$9.2$ & $2.1$&$5.5$ &$4.4$\\
MaskCLIP~\cite{MaskCLIP} & ECCV'22 & &$70.0^\dag$ & $21.7$ & $12.2^\dag$ & $19.8^\dag$ & $13.6$\\
SCLIP~\cite{SCLIP} & ECCV'24 & &$76.9^\dag$ & $32.0$&$12.0^\dag$ &$15.3^\dag$ &$20.5$ \\
ClearCLIP~\cite{ClearCLIP}$^\dag$& ECCV'24&&$80.9$ & $34.9$&$16.2$ &$18.3$ &$22.9$ \\
\hline
MaskCLIP+ & ECCV'22 &\multirow{8}{*}{\makecell[c]{C-USS}}& $70.0^\dag$ & $31.1$ & $12.2^\dag$ & $25.2^\dag$ & $19.5^\dag$ \\
CLIPpy~\cite{CLIPpy} & ICCV'23 & & $54.6$ & - & $13.5$ & $22.3$ & -\\
ReCo~\cite{ReCo} & NeurIPS'22 & & $55.2^\dag$ & $26.2$ & -& $19.3$ & $26.3$\\
ReCo~\cite{ReCo}(w/o. CE) $^\dag$ & NeurIPS'22 & & $54.8$ & $23.1$ & -& $18.2$ & $20.8$\\
CLIP-S4~\cite{CLIP-S4}& CVPR'23 & & $72.0$ & $33.6$ & -& -&- \\
ReCLIP (rec.)~\cite{ReCLIP}& CVPR'24 & & $58.5$ & $25.8$ & $11.1$ & $16.2$ &$14.6$\\
ReCLIP~\cite{ReCLIP}& CVPR'24 & & $75.8$ & $33.8$ & $14.3$ & $19.9$ &$20.3$\\
\rowcolor[gray]{0.9}ReCLIP++ & Ours & &$\bf{85.4}$ & $\bf{36.1}$ & $\bf{16.4}$ & $\bf{26.5}$ &$\bf{23.8}$\\
\bottomrule[1pt]
\end{tabular}
\end{table}

\noindent \textbf{Comparison.}
The comparisons with previous state-of-the-art methods on five benchmarks are demonstrated in Table~\ref{comparison_table}.
From Table~\ref{comparison_table}, we have the following observations:
(1) C-USS methods consider the same setting as our method. We observe that ReCLIP++ outperforms previous state-of-the-art C-USS methods obviously, achieving new state-of-the-arts on all five benchmarks. 
Since ReCo~\cite{ReCo} employs a ``context elimination'' (CE) trick which introduces prior knowledge to assist the training, we also report the results of ReCo by removing this trick (ReCo w/o. CE in Table~\ref{comparison_table}) for a fair comparison. 
Notably, ReCLIP++ outperforms MaskCLIP+ by $15.4$\%, $5.0$\%, $4.2$\%, $1.3$\% and $4.3$\% respectively on the five datasets and outperforms CLIP-S4 by $13.4$\% and $2.5$\% on PASCAL VOC and PASCAL Context respectively.
All these results verify the effectiveness of our ReCLIP++ in rectifying the bias of CLIP to assist unsupervised semantic segmentation
(2) ReCLIP++ outperforms previous state-of-the-art TF-OVSS methods, implying that the bias of CLIP in complex visual understanding tasks cannot be fully rectified by simply modifying its architecture without any training.
(3) TL-OVSS methods consider a different setting and are trained with large-scale image-caption pairs. Strictly speaking, our method cannot be directly compared with those works.
However, the superior performance of ReCLIP++ compared to those typical TL-OVSS methods still demonstrates the effectiveness of ReCLIP++.

%(1) ReCLIP++ outperforms some typical trainable language-guided OVSS methods supervised by large-scale image-caption pairs.
%This demonstrates that CLIP inherently encodes valuable knowledge for complex visual understanding tasks, enabling the mining of such knowledge without the need for annotations.
%(2) ReCLIP++ outperforms previous state-of-the-art TF-OVSS methods, verifying the bias of CLIP in complex visual understanding tasks cannot be fully rectified by simply modifying its architecture without any training.
%(3) ReCLIP++ outperforms previous state-of-the-art C-USS methods obviously, achieving new state-of-the-arts on all the five benchmarks. 
%Since ReCo~\cite{ReCo} employs a ``context elimination'' (CE) trick which introduces prior knowledge to assist the training, we also report the results of ReCo by removing this trick (ReCo w/o. CE in Table~\ref{comparison_table}) for a fair comparison. 
%Notably, ReCLIP++ outperforms MaskCLIP+ by $15.4$\%, $5.0$\%, $4.2$\%, $1.3$\% and $4.3$\% respectively on the five datasets and outperforms CLIP-S4 by $13.4$\% and $2.5$\% on PASCAL VOC and PASCAL Context respectively.
%All these results verify the effectiveness of our ReCLIP++ in rectifying the bias of CLIP to assist unsupervised semantic segmentation.}

\subsection{Comparison with Conference Version}
In Table~\ref{ablation-effectiveness}, we verify the effectiveness of each technical improvement of ReCLIP++ beyond the rectification stage of ReCLIP (rec.)~\cite{ReCLIP} (our conference version):
1) we additionally introduce a decoder that takes the rectified logit map and visual feature of CLIP as input and outputs smoother and more contextual masks (denoted as ``Decoder");
2) we design a new strategy to generate a more accurate label hypothesis for each image (denoted as ``LH"); 
3) we optimize the design of the bias encoding scheme to independently encode the class-preference and space-preference bias (denoted as ``Indep.(Multiplication)").
In Table~\ref{ablation-effectiveness}, we observe that when sequentially adding each technical component, the segmentation performance is consistently improved, which verifies the effectiveness of each technical contribution. 

Finally, as shown in Table~\ref{comparison_table}, with all these technical improvements, ReCLIP++ remarkably outperforms ReCLIP. 
Notably, we observe that even without a distillation, ReCLIP++ still outperforms ReCLIP on all benchmarks, which exhibits a stronger bias rectification capability of ReCLIP++ compared with ReCLIP.
For example, on PASCAL VOC and Cityscapes, ReCLIP++ outperforms ReCLIP by $9.6\%$ and $6.6\%$ respectively.
 
\begin{table}[!t]
\setlength{\tabcolsep}{5.3pt}
\caption{
\textbf{Technical improvements of ReCLIP++ compared with ReCLIP.}
The first line is the results of ReCLIP after the rectification stage, and the following lines are ablation experiments on the technical improvements of ReCLIP++ beyond ReCLIP. 
``LH'' represents Label Hypothesis, while ``Indep.'' represents independent encoding of bias.
Results show that each newly added technical improvement of ReCLIP++ contributes to better results than ReCLIP.
\label{ablation-effectiveness}}
\centering
\renewcommand{\arraystretch}{1.4}
\begin{tabular}{ccccc|cc}
\hline
ReCLIP (rec.) & Decoder & LH & Indep.(Addition) & Indep.(Multiplication) &  VOC & Context \\
\hline
$\checkmark$ &  & & & &$58.5$ & $25.8$\\
 \hline
$\checkmark$ & $\checkmark$ & & & &$75.6$ & $28.3$ \\

{$\checkmark$} & {$\checkmark$} & {$\checkmark$} & & &{$77.3$} &  {$31.4$}\\
 & $\checkmark$ & $\checkmark$ &$\checkmark$ & &{$81.1$} &  {$33.3$}\\
\rowcolor[gray]{0.9} & $\checkmark$ & $\checkmark$ & &$\checkmark$ & $\bf{85.4}$ & $\bf{36.1}$ \\
\hline
\end{tabular}
\end{table}

\subsection{Ablation study}
\label{ablation_study}

\begin{table}[!t]
\setlength{\tabcolsep}{14.5pt}
\renewcommand{\arraystretch}{1.6}
\caption{
\textbf{Ablation on whether Query should be learnable.} 
The $\times$ means fixed while $\checkmark$ means learnable.
\label{ablation-prompt-table}}
\centering
\begin{tabular}{cc|ccc}
\hline
Query & Reference & PASCAL VOC & PASCAL Context & Cityscapes\\
\hline
$\checkmark$ & $\checkmark$ & 11.6 & 3.6 & 2.3\\
\rowcolor[gray]{0.9}$\times$ & $\checkmark$ & \textbf{85.4} & \textbf{36.1}& \textbf{26.5}\\
\hline
\end{tabular}
\end{table}

\noindent \textbf{Should Query be learnable?}
For text inputs $Q$ and $R$, we only make Reference $R$ learnable while keeping Query $Q$ fixed.
In this ablation, we study whether the prompt of Query should also be learnable. 
As shown in Table~\ref{ablation-prompt-table}, we conduct experiments with learnable and fixed Query respectively. 
The numbers show that fixing Query obviously outperforms making Query learnable.
The reason is when we make Query learnable, it may also implicitly capture bias, confusing the following bias rectification operations (\emph{e.g.,} bias logit subtraction).
Therefore, in our framework, we choose to fix $Q$ to make the bias rectification more effective.

\noindent {\textbf{Effect of different fusion strategies in bias logits generation.}}
{In Table~\ref{ablation-effectiveness}, we conduct an ablation comparing two fusion strategies for $W_r$ and $W_p$: addition~(Indep.(Addition)) and matrix multiplication~(Indep.(Multiplication)).
Comparing Line 3 \& 4, we find that even when using the addition fusion strategy of $W_r$ and $W_p$, ReCLIP++ with the independent bias encoding strategy still remarkably outperforms the model without such a strategy.
This demonstrates that our independent bias encoding strategy in ReCLIP++ contributes significantly to the performance improvement.
By comparing Line 4 \& 5, we find that fusing $W_r$ and $W_p$ via matrix multiplication yields better results.
Based on the results above, we adopt matrix multiplication for fusing $W_r$ and $W_p$ in ReCLIP++ to achieve the best results.}

\noindent \textbf{Effect of element-wise subtraction on bias rectification.}
In order to validate the effect of our element-wise subtraction, we conduct experiments in Table~\ref{ablation-minus}.
We compare our subtraction mechanism with an alternative solution:
instead of subtracting the bias logit map $M_b$ from the Query logit map $M_q$, we add $M_b$ to $M_q$. 
Comparisons shown in the table verify the effectiveness of our subtraction way.
As we aim to remove the bias from the original CLIP, we speculate that the subtraction operation may work as a strong prior which regularizes the training to facilitate meaningful and effective bias modeling.

\begin{table}[!t]
\setlength{\tabcolsep}{13pt}
\renewcommand{\arraystretch}{1.6}
\caption{
\textbf{Effectiveness of element-wise subtraction.} 
\textnormal{Results verify the effectiveness of our subtraction way.}
\label{ablation-minus}}
\centering
\begin{tabular}{cc|ccc}
\hline
Subtraction &  Addition & PASCAL VOC & PASCAL Context & Cityscapes\\
\hline
$\times$ & $\checkmark$ & 77.7 & 33.3 &  21.5\\
\rowcolor[gray]{0.9}$\checkmark$ & $\times$ & \textbf{85.4} & \textbf{36.1} & \textbf{26.5}\\
\hline
\end{tabular}
\end{table}

\noindent \textbf{Sensitivity to $t$ and $r$ in image-level multi-label hypothesis generation.}
When generating an image-level multi-label hypothesis as discussed in Sec.~\ref{3.4}, $t$ and $r$ need to be determined. We study the sensitivity of our method to these hyper-parameters.
As shown in Fig.~\ref{hyper-parameter figure}, we observe that the mIoU of our method firstly increases and then decreases as $t$ or $r$ increases, exhibiting a typical bell curve. Such a curve shows the regularization effect of $t$ and $r$ on the training.
By replacing the multi-label hypothesis generation solution in ReCLIP++ with that used in ReCLIP (\emph{i.e.,} ``w/o. crop" in the figure), we obtain results that are inferior to ReCLIP++ within a vast range of $t$ or $r$.
All the results show that the solution adopted in ReCLIP++ is insensitive to the choices of $t$ and $r$ and is superior to that used in ReCLIP.
\begin{figure}
  \centering
  \includegraphics[width=1.0\linewidth]{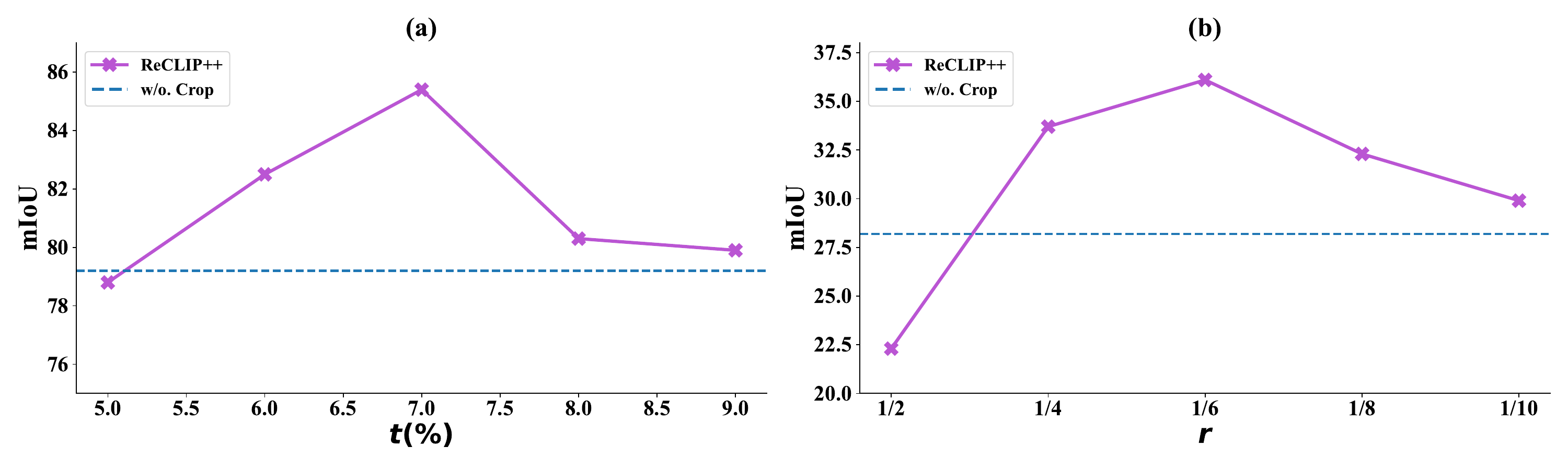}
  \caption{\textbf{Sensitivity to $t$ and $r$ in the image-level multi-label hypothesis generation.}
  (a) Sensitivity to threshold $t$. {In the image-level multi-label hypothesis generation process, we set the window width and height to $r$ times the width and height of the input image.}
  (b) Sensitivity to $r$. {In the image-level multi-label hypothesis generation process, $t$ is the threshold for choosing classes with high frequency within window crops. Results of (a) and (b) verify that ReCLIP++ is insensitive to the choices of $t$ and $r$ and within a vast range of hyperparameter choices, ReCLIP++ remains superior to the solution which employs the multi-label hypothesis strategy of ReCLIP~(denoted as ``w/o. Crop'').}}
  \label{hyper-parameter figure}
\end{figure}

\noindent \textbf{Effect of Fine-tuning CLIP.}
We conduct experiments with fine-tuning the image encoder of CLIP in our framework. The results on PASCAL VOC, PASCAL Context, and Cityscapes are shown in Table~\ref{ablation-finetune}. We find a consistent performance decrease compared with freezing CLIP. 
This is because fine-tuning CLIP on small-scale downstream datasets may harm the image-text alignment in pre-trained CLIP.

\begin{table}[!t]
\setlength{\tabcolsep}{16pt}
\renewcommand{\arraystretch}{1.8}
\caption{
\textbf{Effect of Fine-tuning CLIP.} 
\textnormal{The ``Fine-tune CLIP" means we fine-tune the image encoder of CLIP.}
\label{ablation-finetune}}
\centering
\begin{tabular}{c|ccc}
\hline
Setting & PASCAL VOC & PASCAL Context & Cityscapes\\
\hline
Fine-tune CLIP& 76.5 &  33.7 & 20.4 \\
\rowcolor[gray]{0.9}Frozen CLIP (Ours)& \textbf{85.4} & \textbf{36.1} & \textbf{26.5}\\
\hline
\end{tabular}
\end{table}

\begin{table}[!t]
\setlength{\tabcolsep}{15pt}
\renewcommand{\arraystretch}{1.6}
\caption{
\textbf{Effectiveness of bias rectification.} 
\textnormal{Results further verify that the rectification of both biases contributes to a better segmentation performance.}
\label{ablation-bias}}
\centering
\begin{tabular}{cc|cc}
\hline
Class-preference &  Space-preference & PASCAL VOC & PASCAL Context\\
\hline
$\checkmark$ & $\times$ & 76.4 & 34.1\\
$\times$ & $\checkmark$ & 76.6 & 32.4\\
\rowcolor[gray]{0.9}$\checkmark$ & $\checkmark$ & \textbf{85.4} & \textbf{36.1}\\
\hline
\end{tabular}
\end{table}

\begin{figure}
  \centering
  \includegraphics[width=1.0\linewidth]{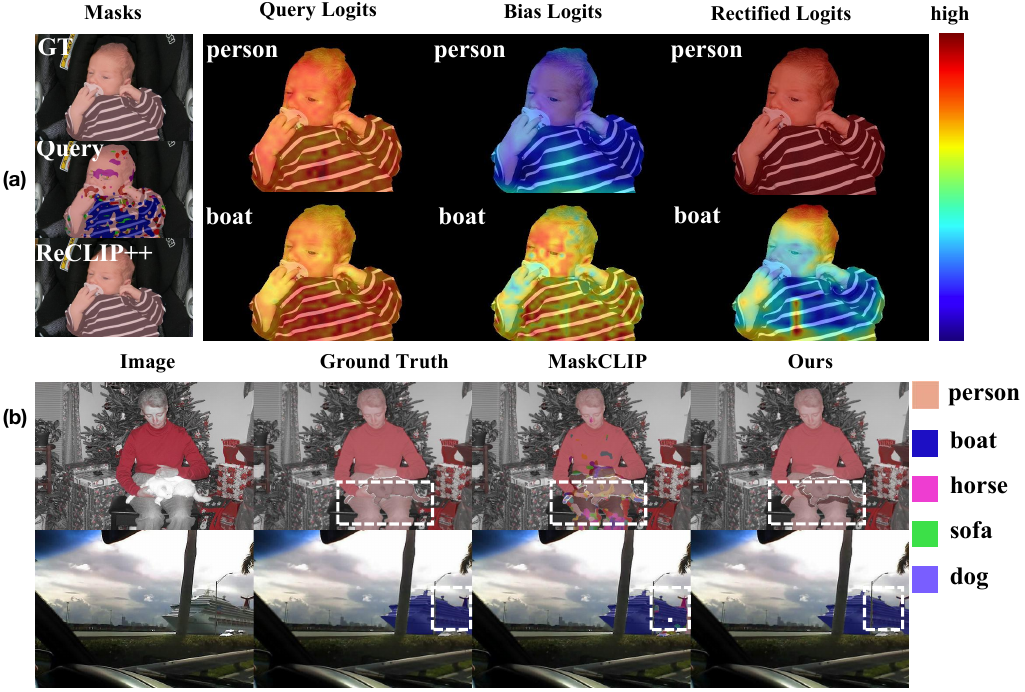}
  \caption{\textbf{Visualization of Bias Rectification.} \textbf{(a) Class-preference bias}: In order to explain how the class-preference bias is explicitly encoded, we show the heatmap of the bias logit map. \textbf{(b) Space-preference bias}: The segmentation within dashed boxes shows the effectiveness of our method on rectifying space-preference bias.}
  \label{visualization_bias_result}
\end{figure}

\noindent {\textbf{Effect of different bias rectification.}}
We conduct experiments to evaluate how the rectification of each bias affects the segmentation performance in Table~\ref{ablation-bias}.
Technically, when testing the rectification effect of the class-preference bias, we perform average pooling on the positional feature $W_p$ along the spatial dimension and expand the pooling result to the original shape of $W_p$.
Similarly, when testing the rectification effect of the space-preference bias, we perform average pooling on the reference text feature $W_r$ along the class dimension and expand the pooling result to the original shape of $W_r$.
Results on PASCAL VOC and PASCAL Context shown in Table~\ref{ablation-bias} further verify that the rectification of each bias contributes to a better segmentation performance.

\noindent {\textbf{Bias Rectification Evaluation.}}
In this ablation, we further evaluate the rectification effect of two biases, both qualitatively and quantitatively.

\noindent\textbf{(1) Qualitative evaluation of bias rectification.}
In Fig.~\ref{visualization_bias_result} (a), we illustrate how we explicitly model and rectify the class-preference bias.
As shown in the first column of Fig.~\ref{visualization_bias_result} (a),
the original CLIP (see ``Query" mask) tends to misclassify part of a ``person" (see the ground-truth mask denoted as ``GT") into a ``boat".
Such a mistake is reflected in the comparison between the logit heatmaps for the ``person" channel and the ``boat" channel:
in the area misclassified as ``boat", the logits for ``boat" are relatively higher than those for ``person".
In contrast, for the bias logit map, the person-channel logits are quite low, while the boat-channel logits are generally very high, especially for the misclassified area.
Consequently, via the proposed element-wise logit subtraction operation, we obtain the rectified logit map (see the last column), where the boat channel is largely suppressed. Finally, we obtain a much better mask (see ``ReCLIP++" in the first column).
In Fig.~\ref{visualization_bias_result} (b), we aim to show that the space-preference bias is effectively modeled and rectified. 
By comparing the results of the original CLIP~(see the dashed boxes of ``Query" mask) and ReCLIP++~(see the dashed boxes of the last two columns), we find that the segmentation performance in the boundary areas is largely improved.

\noindent\textbf{(2) Quantitative evaluation of bias rectification.}
We further design two metrics to quantitatively evaluate the rectification results with respect to class-preference and space-preference bias respectively.
For class-preference bias, 
we first compute the confusion matrix for all classes in the dataset.
Given a class $i$, we take the $i$-th row of the confusion matrix.
We perform a subtraction, between the $i$-th element and the maximum value among other positions of this row, as the score for class $i$.
Then we average the scores for all classes. 
A higher average score reflects lower bias and better rectification results.
For space-preference bias, we compute the slopes of the curve between the distance and mIoU at different distances (see Fig.~\ref{introduction_fig} and Appendix for how to draw the curve) which roughly denote $\triangle \text{mIoU} / \triangle \text{Distance}$. We average the slopes at different distances.
Obviously, a lower space-preference bias or a better rectification of space-preference bias should bring a higher score.

We show results on PASCAL Context for class-preference bias,
and results on PASCAL VOC for space-preference bias, as there exhibits more obvious class-preference bias on PASCAL Context while more obvious space-preference bias on PASCAL VOC.
According to results shown in Table~\ref{ablation_bias_rectification}, 
both ReCLIP (rec.) and ReCLIP++ can obviously rectify the class-preference and space-preference bias, compared with MaskCLIP which represents the original segmentation ability of CLIP.
Compared with ReCLIP (rec.), ReCLIP++ exhibits stronger rectification results on both biases.

\begin{table}[h]
\setlength{\tabcolsep}{11.5pt}
\renewcommand{\arraystretch}{1.6}
\caption{
\textbf{Quantitative Evaluation of Bias Rectification.}
\textnormal{We design two metrics to evaluate the rectification effect on class-preference and space-preference bias respectively. Higher scores indicate lower bias or stronger rectification effect.}
\label{ablation_bias_rectification}}
\centering
\begin{tabular}{c|cc}
\hline
Method & Class-preference Rectification & Space-preference Rectification\\
\hline
MaskCLIP~\cite{MaskCLIP} & 0.29 & -0.50\\
MaskCLIP+~\cite{MaskCLIP} & 0.31 & -0.48\\
ReCLIP~(rec.)~\cite{ReCLIP} & 0.33 & -0.41\\
\rowcolor[gray]{0.9}ReCLIP++&\textbf{0.40}& \textbf{0.05}\\
ReCLIP++ \& distill & 0.37 & -0.17\\
\hline
\end{tabular}
\end{table}

% \begin{table}[!t]
% \setlength{\tabcolsep}{20pt}
% \renewcommand{\arraystretch}{1.6}
% \caption{
% \textbf{Effect of components for distillation.}
% Results show that it is better to inherit and freeze the components, including query segmentation head, bias logits and the Rectified Mask Generation Module, from the rectification stage than learning the original type of classification head.
% \label{sup_ablation_distillation}}
% \centering
% \begin{tabular}{cccc|c}
% \hline
% Ori. Head & Query Head& Bias Logits & RMG & mIoU\\
% \hline
% $\checkmark$ & $\times$ & $\times$ & $\times$ & $77.3$\\
% $\times$ & $\checkmark$ & $\times$ & $\times$ & $57.2$\\
% $\times$& $\checkmark$ & $\checkmark$& $\times$ & $60.5$\\
% \rowcolor[gray]{0.9}$\times$& $\checkmark$ & $\checkmark$ & $\checkmark$ & $\bf{80.0}$\\
% \hline
% \end{tabular}
% \end{table}

\noindent {{\textbf{Effect of different inputs of mask decoder $\mathcal{F}_{dec}$}.}} 
{In this ablation, we design an ablation on the effect of different inputs of the mask decoder $\mathcal{F}_{dec}$ in Table~\ref{ablation_decoder_input}.
Line 1 refers to using only the rectified logit map $M$ as the input of the decoder, while Line 2 refers to ReCLIP++, which uses both the feature $Z$ and the rectified logit map $M$ as inputs. 
Results on PASCAL VOC~\cite{PASCAL_VOC} and PASCAL Context~\cite{PASCAL_Context} show that ReCLIP++ (Line 2) outperforms the strategy that uses only $M$ as the input of $\mathcal{F}_{dec}$ (Line 1).
Besides, we quantitatively evaluate both class-preference and space-preference bias rectification effects using our designed metrics described in Section~\ref{ablation_study} (``Bias Rectification Evaluation (2)'').
Results in Table~\ref{ablation_decoder_input} show that including $Z$ as the input of $\mathcal{F}_{dec}$ further improves bias rectification compared to using $M$ alone.}

\begin{table}[h]
\setlength{\tabcolsep}{6.2pt}
\renewcommand{\arraystretch}{1.8}
\centering
\caption{
\textbf{{Ablations on inputs of the decoder.}}
\textnormal{{We compare these two strategies by both performance and bias rectification effect. The performance is represented by mIoUs on PASCAL VOC and PASCAL Context and the bias rectification is represented by our designed metrics. A higher number of ``Class-Preference Rectification'' or ``Space-Preference Rectification'' means a stronger bias rectification effect.}}
\label{ablation_decoder_input}}
\begin{tabular}{cc|cccc}
\hline
{$Z$} & {$M$} & {VOC} & {Context} & {Class-Preference Rectification} & {Space-Preference Rectification}\\
\hline
 & {$\checkmark$} & {$73.6$} & {$28.0$} & {$0.34$} & {$-0.35$} \\
\rowcolor[gray]{0.9} 
{$\checkmark$}&{$\checkmark$}& {\textbf{85.4}} & {\textbf{36.1}} & {\textbf{0.40}} & {\textbf{0.05}}\\
\hline
\end{tabular}
\end{table}

\noindent {\textbf{Why removing the distillation stage?}}
As discussed in Sec.~\ref{3.4}, in ReCLIP++, we don't employ the distillation stage as adopted in ReCLIP. 
We observe that employing an additional distillation stage after ReCLIP++ leads to inferior results, as shown in Table~\ref{distillation_table}.
Further, we empirically study the effect of distillation on bias rectification.
As shown in Table~\ref{ablation_bias_rectification}, we observe that instead of mitigating the bias, further distillation after ReCLIP++ may enlarge the bias, which has a negative effect on the final segmentation performance.
This phenomenon explains why the distillation stage is no longer necessary in ReCLIP++.
In ReCLIP++, thanks to $\mathcal{F}_{dec}$, the rectified mask of ReCLIP++ is much smoother and contextual than that of ReCLIP, rendering the negative effect of distillation outweighs the positive effect of distillation and leading to a decrease of segmentation performance.

%In ReCLIP~\cite{ReCLIP}, an additional distillation stage is employed to distill the knowledge of rectified CLIP into a specifically designed segmentation network like DeepLab-v2~\cite{Deeplabv2}.
%However, in ReCLIP++, we don't need such a distillation stage.
%Empirically, we observe that such a distillation process has two opposing effects.
%On the one hand, it may benefit the smoothness of the predicated mask, which has a positive effect on the segmentation performance.
%On the other hand, instead of mitigating the bias, it may enlarge the bias to an extent, which has a negative effect on the segmentation performance.
%As shown in Table~\ref{distillation_table}, we distill the knowledge of ReCLIP++ into DeepLab-v2~\cite{Deeplabv2}.
%The distillation is similar to that in ReCLIP and the rectified mask is generated as Eq.~(2) to Eq.~(4) in Sec.~\ref{sec3}.
%Furthermore, according to Table~\ref{ablation_bias_rectification}, we observe that distillation (denoted as ``ReCLIP++ \& distill'') harms the rectification of both biases compared with ReCLIP++.
%The reason is that, with $\mathcal{F}_{dec}$, the rectified mask of ReCLIP++ is much smoother and contextual than that in ReCLIP, rendering the negative effect of distillation outweighs the positive effect of distillation and leading to a decrease in segmentation performance in Table~\ref{distillation_table}.
%Thus, we choose not to employ the distillation in ReCLIP++ to make the training more simplified and achieve even better segmentation performance.}

\begin{table}[!t]
\setlength{\tabcolsep}{21pt}
\renewcommand{\arraystretch}{2}
\caption{
\textbf{Distillation results of ReCLIP++.} Results show that distillation cannot bring consistent improvement on all benchmarks anymore.
\label{distillation_table}}
\centering
\begin{tabular}{c|ccc}
\hline
Method & PASCAL VOC & PASCAL Context & ADE20K\\
\hline
ReCLIP++ & $\bf{85.4}$ & $\bf{36.1}$ & $\bf{16.4}$\\
Distillation & $80.0$ & $34.5$ & $15.8$ \\
\hline
\end{tabular}
\end{table}

\noindent \textbf{Qualitative Results.}
We visualize our segmentation results in Fig.~\ref{visualization_result}. It can be observed that there exist apparent space-preference bias and class-preference bias in the segmentation results of the original CLIP (MaskCLIP). With ReCLIP, both biases can be partially removed. 
Moreover, ReCLIP++ exhibits stronger bias rectification capability compared with ReCLIP, yielding the best segmentation results. 
\label{qualitative results}

\begin{figure*}
  \centering
  \includegraphics[width=1.0\linewidth]{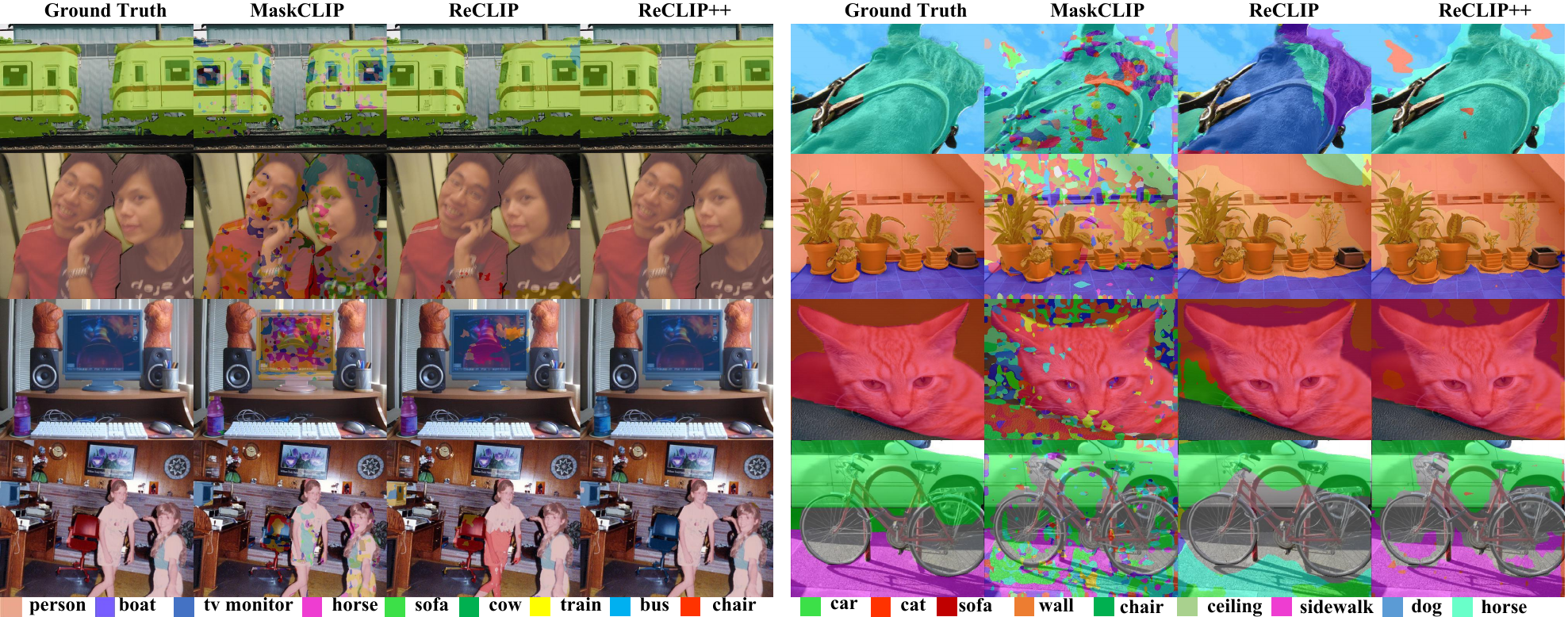}
  \caption{\textbf{Qualitative Results}: We visualize segmentation results on both PASCAL VOC (left) and PASCAL Context (right). From the visualization, we observe that ReCLIP++ outperforms both MaskCLIP and ReCLIP obviously by rectifying both class-preference bias and space-preference bias.}
  \label{visualization_result}
\end{figure*}

\section{Conclusion}\label{sec5}
In this paper, we propose a new framework for language-guided unsupervised semantic segmentation. 
We observe the bias, including space-preference bias and class-preference bias, exists in CLIP when directly applying CLIP to segmentation task. 
We propose using Reference text feature to encode class-preference bias and projecting positional embedding to encode space-preference bias independently, and then manage to combine them into a bias logit map by matrix multiplication.
By a simple element-wise logit subtraction mechanism, we rectify the bias of CLIP.
Then a contrastive loss is imposed to make the bias rectification meaningful and effective.
Extensive experiments demonstrate that our method achieves superior segmentation performance compared with previous state-of-the-arts. We hope our work may inspire future research to investigate how to better adapt CLIP to complex visual understanding tasks.

\newpage

% \section*{Data Availability Statement}
% All data that support the findings of this study are openly available.
% PASCAL VOC 2012 is available at \url{http://host.robots.ox.ac.uk/pascal/VOC/} and PASCAL Context is available at \url{https://cs.stanford.edu/~roozbeh/pascal-context/}.
% ADE20K can be downloaded from \url{https://ade20k.csail.mit.edu/}, while Cityscapes can be downloaded from \url{https://www.cityscapes-dataset.com/}.
% Homepage of COCO Stuff is \url{https://github.com/nightrome/cocostuff}.

\bibliography{sn-bibliography}% common bib file

\end{document}